# Is Luke the Author of a Gospel and the Acts of the Apostles?

Jacques Savoy

Computer Science Department, University of Neuchâtel,
Rue Emile Argand 11, 2000 Neuchâtel, Switzerland

Jacques.Savoy@unine.ch

**Abstract**

According to Christian tradition, Luke is credited with authoring a Gospel and the Acts of the Apostles, even if his name does not appear in either book, both originally written in Koine Greek. Several biblical scholars assume that both texts were written by a common author, while others deduce the presence of two authors. Different studies have been found to support either finding, some based on qualitative evaluation, while a few others consider the occurrence frequency differences between the two books. To propose an enhanced quantitative analysis, this study is grounded on two recent authorship attribution models. The Burrows' Delta, applied with eleven different feature sizes, demonstrates common authorship. An author verification model confirms this finding. The following experiments consider several stylistic representations, feature sizes, and distance functions to confirm that Luke is the true author of both books.

## 1. Introduction

In Christian tradition, Luke is known as the author of the Gospel of Luke and the Acts of the Apostles (Carson et al. 2005; Holladay 2016). These two works represent around 28% of the New Testament and are a well-known authorship attribution problem (Love 2002; Savoy 2020; Kreuz 2023). Several arguments have been put forward to assume this common authorship (Cadbury 1927; Burkett 2019; Gregory & Rowe 2010). Firstly, both books begin with a dedication to Theophilus, as shown in the Acts:

> " 1 The former treatise have I made, O Theophilus, of all that Jesus began both to do and teach,
> 2 until the day in which he was taken up, after that he through the Holy Ghost had given commandments unto the apostles whom he had chosen:" (Acts, 1:1-2)

And in Luke's Gospel:

> " 1 Forasmuch as many have taken in hand to set forth in order a declaration of those things which are most surely believed among us,
> 2 even as they delivered them unto us, which from the beginning were eyewitnesses, and ministers of the word;

3 it seemed good to me also, having had perfect understanding of all things from the very first, to write unto thee in order, most excellent Theophilus," (Luke 1:1-3)

This common dedication is viewed as an indication that the author is a well-educated member of Hellenistic society (Tyson 2011). As a second piece of evidence, one can see a continuum between these two books: Luke's Gospel ends with Jesus ascending to Heaven, while the Acts begin with the disciples waiting in Jerusalem for the Holy Spirit. Moreover, at the beginning of the Acts, the author writes that he already wrote a previous book. Additionally, according to Christian tradition, the Luke-Acts[1] notation highlights this close proximity between the two books (Cadbury 1927).

Third, a topical relationship clearly appears between the contents of both books in their justification of preaching not to the Jews, but to the Gentiles[2] (Luke 4:16-30; 7:9; 8:19-21; 14:15-24). To emphasise this aspect, Luke's Gospel ends with "to all the nations [Gentiles] beginning from Jerusalem" (Luke 24:47), (Burkett 2019). To Luke, the Jews rejected the message, requiring the convincing of the Gentiles.

Fourth, the two books emphasise the role played by the Holy Spirit, the presence of several sermons spoken for different audiences, the establishment of communities outside Judea (e.g. Antioch), and the innocence of Christians (they had not formed a subversive movement against the Roman administration). To conclude, many scholars emphasise the fact that the same pen could be behind both books:

> "Luke and Acts were written by the same person, whether Luke the physician or not, whether a Jew or a Gentile, whether from Antioch or Ephesus (or somewhere else)." (Parsons & Pervo 1993)

Besides the question of authorship, additional open questions in relation to the books exist such as the narrative, canonical, genre or theological unity[3] between Luke-Acts (Parsons & Pervo 1993; Gregory & Rowe 2010). The answer to each of these inquiries might be an argument in favour of or against common authorship.

> ".. the fact that Luke and Acts were written by the same author (accepted by virtually all scholars) does not necessarily imply that the two books were written together, nor that they should be considered as the two parts of the same work." (Bovon, 2010, p. vii)

However, some scholars raise doubts as to the presence of a single author (Clark 1933; Argyle 1974; Dawsey 1989; Walters 2009; Mealand 2016). These doubts are usually based on the occurrence frequency differences of some terms between the two books. Moreover, when comparing the Gospel to the Acts, the styles are viewed as distinct from each other.

---

[1] Traditionally, scholars denote both books as "Luke-Acts" to emphasise the unity of the Gospel and Acts, with a hyphen signaling the strong connection.

[2] Persons who are not Jewish but interested in Judaism (also called God-fearers).

[3] With the underlying questions: Do these two books really form a continuum of the same story? Why don't they appear in sequence in the canon? Do they correspond to the same text genre? Do they really expose the same theological principles?

Using recent stylometric models, this study evaluates whether or not the same author is behind Luke-Acts. We do not assume that the author's name is Luke, a name corresponding to an important figure and certainly given to provide authoritative evidence and wider acceptance of both texts (Aland 1961; Argyle 1974). For the most part, however, we suppose that a single and common author was responsible for writing Luke-Acts.

The rest of this paper is organised as follows. The next section presents the state of the art while Section 3 describes the corpora used in our experiments. Section 4 exposes some experiments based on the occurrence frequency differences when considering a selection of words. Section 5 depicts the results achieved when applying the Burrows' Delta model to the problem of Luke-Acts. Section 6 presents the outcomes obtained by a recent attribution model used in conjunction with several stylistic representations, features sizes, and similarity approaches. Section 7 shows some stylistic patterns that are more associated to Luke's style. The last section reports the main findings of this study.

## 2. State of the Art

As is the case for other text categorisation tasks (Sebastiani 2002), an effective authorship attribution model (Savoy 2020) must represent each text according to a set of features which closely reflect the author's style. To achieve this, a first family of methods suggests representing each document based on the most frequent word-types (MFW) or according to a predefined set of functional words (e.g. articles, prepositions, pronouns, conjunctions, auxiliary verb forms) considered as independent of the topics. Various distance (or similarity) functions have been proposed to then compare two text surrogates. Following this strategy, Burrows' Delta (Burrows 2002) suggests using the top $m$ most frequent word-tokens (with $m = 50$ to 1,000), while Zhao & Zobel (2007) propose a predefined set of 363 English words. In this latter approach, the Kullback-Leibler divergence was chosen to compare two representations. On the other hand, the Manhattan distance was proposed by Burrows (2002) based on the Z score of the selected terms. With Labbé's method (Labbé 2007), the entire vocabulary is used and a variant of the Manhattan distance is suggested. This approach was found to be effective for author profiling (Kocher & Savoy 2017).

As a second paradigm, multivariate analysis can be applied to project each document's surrogate into a reduced space under the assumption that texts written by the same author will appear close together. Some of the main approaches applicable here are principal component analysis (PCA) (Binongo & Smith 1999; Craig & Kinney 2009), hierarchical clustering (Labbé 2007; Tuzzi & Cortelazzo 2018), or discriminant analysis (Jockers & Witten 2010). In terms of stylistic features, these approaches tend to employ the top 50 to 500 most frequent words, as well as some Part-Of-Speech (POS) information.

A third family of possible methods of machine learning approaches has been proposed (James et al. 2021) such as, for example, decision trees, *k*-nearest neighbours, neural networks, random forests, logistic regression and support vector machines (SVM). Depending on the test collection and the representation strategy, one of these strategies tends to perform better than the others, but it is difficult to specify beforehand which one (Ikae & Savoy 2022).

While words seem to be a natural way to generate a stylistic surrogate, other studies have suggested using the letter occurrence frequencies (Kjell 1994) or the distribution of short sequences of letters (character *n*-grams) (Stamatatos et al. 2015). As demonstrated by Kešelj et al. (2003), such a representation could produce high performance levels. However, when adopting such a strategy, the final decision is more difficult to explain to the user (e.g. what is the stylistic meaning of a frequent use of "ui"?). The fingerprint of an author can also be identified by the POS tags distribution or short sequences of such tags (Kocher & Savoy, 2018). Such text representations do not usually produce the best performance levels but can be used to provide useful complementary information (Zheng et al. 2006). Moreover, this approach assumes the use of an effective and efficient POS tagger[4].

Finally, to solve the verification question (determining whether or not a given author did in fact write a given text), some modifications of these strategies must be applied. In this context, the training sample contains texts written by a single author who might also be the writer of the query document. To achieve this, the disputed text (denoted Q and assumed to be written by A) can be processed as a whole or as a sequence of *c* chunks (e.g. each composed of 5,000 tokens). The result obtained by these *c* subparts of Q determines the final answer (Koppel et al. 2007). As a variant, a set of other possible writers called *impostors* (with a text sample for each of them) can be included. A set of binary classifiers is trained to learn models for A vs. not-A, B vs. not-B, etc. The *c* chunks of the query text are then classified according to the learned models, and if a preponderance of chunks is classified as A, then we conclude that A is the real author (Koppel & Winter 2014). Finally, Koppel & Seidman (2018) suggest iterating by selecting only a fraction of the entire possible stylistic features. The selection can be iterated with a different randomly chosen subset of features. The final selection of these numerous random features depends on the majority of the cases found similar to the author A or one of the others.

## 3. Evaluation Corpus

As for other Christian writings, the original Luke and Acts were written in Koine Greek[5], a *lingua franca* of the time. For this experiment, a corpus of 190 chapters was downloaded from the website *Bibelwissenschaft.de* (Nestle-Aland *Novum Testamentum Graece* edition). The texts were then preprocessed to remove additional elements (e.g. verse numbers, running titles). The English examples were extracted from the King James version, which was downloaded from the same website.

Various information about these books is reported in Table 1 (given in the canonical order of the New Testament). In the second column, the commonly attributed author's name is provided. For John and Luke, a quote has been added from the Acts, as well as from the Book of Revelation (or Apocalypse of John) to signal a possible different writer than Luke or John. The next column indicates the approximate year of writing, though opinion varies regarding the dates (Robinson 1976). However, according to Pervo (2006), the Acts could have been written considerably later than 85-90. This assumption renders the presence of a single author for both Luke's Gospel and the Acts more problematic.

---

[4] As for all natural languages, the correct assignment of a POS tag could be ambiguous. The word 'still', for example, could be an adverb, noun, adjective, or even a verb.

[5] To simplify the notation, the term "Greek" means Koine (or Ancient) Greek in the rest of this study.

In addition, the number of chapters in each book is reported. In the last two columns, the number of tokens and word-types are reported.

As indicated in Table 1, each Gospel and the Acts contains more than 10,000 tokens. It is worth knowing that text length is an important factor for a reliable outcome to be provided. When employing texts with a mean length clearly below 5,000 words, reliable authorship attribution is rather difficult to achieve (Eder 2015; Savoy 2018). The four Pauline letters correspond to those found to be genuine by Baur (1845), a finding confirmed by a recent stylometry study (Savoy 2019).

| Book | Author | Year | Chapters | Tokens | Vocabulary |
|---|---|---|---|---|---|
| Matthew | Matthew | 80-85 | 28 | 18,346 | 4,188 |
| Mark | Mark | 65-70 | 16 | 11,304 | 3,002 |
| Luke | Luke | 80-85 | 24 | 19,482 | 4,850 |
| John | John | 90-95 | 21 | 15,635 | 2,802 |
| Acts | 'Luke' | 85-90 | 28 | 18,450 | 4,837 |
| 1 Corinthians | Paul | 53-54 | 16 | 6,832 | 2,056 |
| 2 Corinthians | Paul | ~55 | 13 | 4,478 | 1,488 |
| Romans | Paul | 55-57 | 16 | 7,116 | 2,088 |
| Galatians | Paul | 55-60 | 6 | 2,231 | 903 |
| Revelation | 'John' | 95-99 | 22 | 9,851 | 2,213 |

Table 1: List of the selected books from the New Testament (canonical order)

Unlike more recent documents, an oral tradition may stand behind some of these writings. Once transcribed into Greek, textual variations were introduced through various mechanisms. Some differences are due to alternative ways to reproduce a text. A copyist can work alone, or a reader can dictate the manuscript to a group of them all at once. In the latter, a larger number of errors could appear in the copies (due to the imperfect graphophonics relationship).

As a second source of variation, the copyist could skip a phrase or a line when reproducing the exemplar manuscript due to poor working conditions such as light, lassitude, temperature, etc. Third, the copyist could add a few words or an explanatory gloss. For example, where the name "Jesus" appears in the original text, the monk could write it as "Lord Jesus," "Jesus Christ" or "Son of God". Similarly, the term "Spirit" could be rewritten as "Holy Spirit" (Boismard & Lamouille 1980).

Fourth, an explanatory clause could be inserted where the original passage is viewed as too concise. For example, one finds the following three verses in the Acts though some interpreters consider verse 37 to be an addition. The text still flows without this verse.

> "36 And as they went on their way, they came unto a certain water: and the eunuch said, See, here is water; what doth hinder me to be baptized?
> 37 *And Philip said, If thou believest with all thine heart, thou mayest. And he answered and said, I believe that Jesus Christ is the Son of God*. [italics added]
> 38 And he commanded the chariot to stand still: and they went down both into the water, both Philip and the eunuch; and he baptized him."

(Acts 8:36-38)

With a detailed inspection, some of these variations can be detected by comparing several sources of the same text (e.g. Greek, Latin, Syriac, Coptic, Armenian, Ethiopic, or Georgian version) (Boismard & Lamouille 1984).

In our experiments, Matthew and Mark's gospels are not always used. With the third one written by Luke, they form the synoptic set which shares similar wordings of many passages (Burkett 2019). An example is depicted in Table 2 and others are shown in the Appendix. With such strong similarities of passages, it is not reasonable to consider that Matthew or Mark's gospels can be employed to propose distinct writing styles. When analysing Luke's Gospel, one can estimate that it has around 41% in common with the other synoptic evangelists' texts. In addition, Luke's Gospel appears to have 23% of its contents in common with Matthew's, while 35% is unique (Goodacre 2001; Boismard & Lamouille 1980).

| Mark | Matthew | Luke |
|---|---|---|
| 8,35 For whosoever will save his life shall lose it; but whosoever shall lose his life for my sake and the gospel's, the same shall save it. | 16,25 For whosoever will save his life shall lose it: and whosoever will lose his life for my sake shall find it. | 9,24 For whosoever will save his life shall lose it: but whosoever will lose his life for my sake, the same shall save it. |
| 8,36 For what shall it profit a man, if he shall gain the whole world, and lose his own soul? | 16,26 For what is a man profited, if he shall gain the whole world, and lose his own soul? or what shall a man give in exchange for his soul? | 9,25 For what is a man advantaged, if he gain the whole world, and lose himself, or be cast away? |

Table 2: An example of a strong similarity between the synoptic Gospels (Mark, Matthew, and Luke)

Finally, stylometric analysis must take into account some of the linguistic differences between English and Greek. As for other languages, the most frequent tokens in Greek correspond to determiners (e.g. ὁ, οἱ, ἡ, τὸ (the)), prepositions (πρὸς (to)), conjunctions (καὶ (and, the most frequent word in our corpus)), pronouns (ἐγὼ (I), μέ (me), σύ (you), αὐτοῦ (it, him)) or modal verb forms (ἐστι, ἐστιν (is)). However, Greek has a larger number of possible forms for functional terms. In fact, this language has three genders (masculine, feminine, neutral), three numbers (singular, dual, and plural), and five grammatical cases (nominative, vocative, accusative, genitive, and dative). The resulting morphology must therefore be viewed as more complex. For example, there is no single translation of the definite determiner 'the' but rather seventeen possible words (e.g. ὁ (masc. sing. nom.), οἱ (masc. plur. nom.), ἡ (femi. sing. nom.), τὸ (neut. sing. nom.), etc.). Finally, the word order is relatively free compared to English.

## 4. Frequencies Considerations

One century ago, a first analysis suggested that Luke[6] and Acts share a common authorship (Cadbury 1927). During the last fifty years, several biblical studies have been published in favour of or against this assertion. Different arguments have been put forward based on similar theologies, the similarity in style, or on the canonical order of presentation of these two books (Gregory & Rowe 2010; Burkett 2019). Those approaches are, however, mainly based on qualitative analysis and subjective judgments.

Few studies have considered quantitative information, which takes into account the occurrence frequencies of selected words to support or weaken the case for common authorship (Argyle 1974; Beck 1977; Dawsey 1989). For example, the term *ἐγένετο* (it came) occurs 69 times in Luke, but only 54 times in the Acts (see Table 3). As the two books have approximately the same size (Luke: 19,482 tokens, Acts: 18,450 tokens), these examples are given in absolute frequencies. Such cases challenge the hypothesis that Luke and Acts are by the same author.

To further justify the presence of two distinct authors, Clark (1933) and Argyle (1974) cite the conjunction *τε* (and), which appears only nine times in the Gospel but 151 in the Acts. As a third example, one can mention the word *οὖν* (therefore) (33 in Luke, 61 times in the Acts). The noun *μαθητὰς* (disciple) occurs 11 times in Luke followed by a personal genitive, but this word appears seven times in the Acts without a specific pattern (Dawsey 1989). To these examples, one can add the largest differences in occurrence frequencies emerging with the terms *καὶ* (and) (1,466 in Luke, 1,094 in the Acts), *εἶπεν* (he said) (229 in Luke, 71 in the Acts), or *εἰς* (to, into) (225 in Luke, 302 in the Acts). Thus, can we assume that Luke-Acts is authored by the same person?

| Word / Size | Luke 19,482 | Acts 18,450 | Lower Bound | Upper Bound |
|---|---|---|---|---|
| τε (and) | 9 | 151 | 70 | 95 |
| οὖν (therefore) | 33 | 61 | 39 | 58 |
| εἰς (to) | 225 | 302 | 248 | 293 |
| καὶ (and) | 1,466 | 1,094 | 1,197 | 1,293 |
| εἶπεν (he said) | 229 | 71 | 129 | 163 |
| ἐγένετο (it came) | 69 | 54 | 49 | 71 |
| μαθητὰς (disciple) | 11 | 7 | 5 | 13 |

Table 3: Occurrence frequencies of some terms in the Gospel and the Acts

To answer the question of whether or not a word occurs with the same frequency in two distinct samples, Muller (1992) suggests comparing the number of occurrences in the first sample and the whole corpus. For example, knowing that the word *τε* appears 9 times in Luke's Gospel compared to 160 occurrences in the entire corpus (see Table 3), can we infer that *τε* is employed with a similar frequency in the second book, the Acts of the Apostles?

[6] Instead of specifying 'Luke's Gospel', the term 'Luke' will be used to indicate this Gospel.

The test procedure is the following. The size (number of tokens) of the entire corpus is indicated by the variable $n_0$ (= 19,482 + 18,450 = 37,932) while the sample in Luke is denoted $n_1$ (= 19,482) (see Table 3). For the selected term (e.g. τε), one can count its number of occurrences in the entire corpus (value denoted $tf_0$, or 9 + 151 = 160) and its absolute frequency in the Gospel (denoted by $tf_1$, or 9). Assuming that the frequency for this term in the Gospel is the same as for the Luke-Acts (hypothesis $H_0$), one can estimate the probability that $t$ = 'τε' appears $tf_1$ times in the Gospel according to a hypergeometric law (Baayen 2008) described by the following equation:

$$\mathrm{p}(t = tf_1) = \binom{tf_0}{tf_1} \cdot \binom{n_0 - tf_0}{n_1 - tf_1} \Big/ \binom{n_0}{n_1} \tag{1}$$

Estimating only the probability of a single frequency is not fully pertinent and would be very small. One can admit that some natural variability does exist; one author might employ this or that word a little bit more or less, without that indicating a statistically significant deviation. To define the thresholds determining the limits of this variability, one can compute the cumulative distribution of the occurrence frequency of the term $t$ in the underlying corpus as indicated in Equation 2, where $f$ represents the absolute frequency ($f$ = 0, 1, 2, …, $tf_1$) of the chosen word $t$ in Luke :

$$C_t = \mathrm{p}(t \leq tf_1) = \sum_{f=0}^{f=tf_1} \mathrm{p}(t = f) \tag{2}$$

The maximum frequency is $tf_0$, the number of occurrences in the whole corpus. To apply the statistical test, one can define the lower and upper frequency limits (confidence interval) for which the cumulative probability distribution reaches a specified threshold, denoted $\alpha$ and $1-\alpha$ (e.g. $\alpha$ = 5%, 1%, or 0.5%).

For example, according to the null hypothesis $H_0$ and specifying that $\alpha$ = 5% (two-tails), one would expect to observe between 70 and 95 occurrences of *τε* in Luke. As depicted in Table 3, the Gospel contains the word *τε* only nine times, too few occurrences to be inside the confidence interval [70, 95]. Two additional examples (*οὖν* and *εἰς*) are given in Table 3; for these three rows, Luke presents a number of occurrences that is too low according to the null hypothesis.

In the bottom part of Table 3, four examples are depicted for which the Acts present a lower frequency of occurrences. For two of them (namely, *καὶ* and *εἶπεν*), the Acts present a value that is too small, falling outside the confidence interval. For the last two (*ἐγένετο* and *μαθητὰς*), even if the frequency occurrences is lower, the value is still inside the confidence interval.

Based on these results, can we conclude that Luke-Acts is authored by two persons? In order to have a better understanding of such differences, we repeated this experiment with another author, Charles Dickens, and two of his books. First, we chose the novel *Oliver Twist* (first published in 1838, 150,603 tokens) and compared it to *The Mystery of Edwin Drood* (published in 1870, 91,027 tokens).

As shown in Table 4, four terms[7] have been selected as having a clearly higher frequency in one of these two books. For example, the pronoun 'I' presents a lower number of occurrence in *Oliver* compared to *Edwin*. This value is outside the confidence interval. A similar conclusion can be drawn with the auxiliary verb 'is'. On the other end, the words 'the' and 'was' present a lower frequency in *Edwin*, a value outside the expected interval. Can we conclude that these two books have not been written by the same author? Of course not. The stylistic fingerprint of a given novelist cannot be based on a single or a few cases. In addition, we must treat with caution all statistical tests based on a few word occurrence frequencies. In text applications, it is known that each possible probability distribution (e.g. Gaussian, negative binomial, hypergeometric, Poisson, …) does not closely reflect the occurrence distribution for all words (Baayen 2008).

| | Oliver | Edwin | Lower | Upper |
|---|---|---|---|---|
| Word / Size | 150,603 | 91,027 | Bound | Bound |
| I | 1,191 | 1,347 | 1,534 | 1,629 |
| is | 665 | 871 | 920 | 994 |
| the | 9,419 | 4,247 | 5,041 | 5,256 |
| was | 1,777 | 637 | 863 | 956 |

Table 4: Occurrence frequencies of some terms in two novels written by Dickens

## 5. Burrows' Delta

The results of our previous experiments do not amount to an effective method for solving our question of authorship attribution. To begin a deeper quantitative approach, we assume that Luke-Acts has been authored by the same writer, and the following experiments will confirm or reject this hypothesis. As a first approach, the Burrows' Delta (Burrows 2002) model has been selected.

The Delta approach takes into account the most frequent words to ascertain the stylistic markers of a text. One is free to specify the desired number of selected words (value denoted *m*), but the most common values vary between 50 and 500 (Hoover 2004). The underlying idea is to take into account frequent words that carry little or no meaning and are used unconsciously by the author. The majority of them are functional terms (articles, prepositions, pronouns, conjunctions, auxiliary verbs). Moreover, to focus more on the written style, it is important to consider texts extracted from the same text genre (e.g. novels, poems), same time period, and comprised of more than 2,000 to 2,500 letters.

A weight is assigned to each selected term $t_{ij}$, to reflect its importance in a given text $T_j$. This weight, denoted Z score($t_{ij}$), corresponds to the difference between its relative frequency in $T_j$ ($rtf_{ij}$) and the average ($\overline{rtf_i}$) over all texts in the corpus. In addition, to account for the underlying variability, each difference is divided by its standard deviation ($s_i$).

[7] These words belong to the functional set and are less related to the topics of the novels. For example, choosing the names 'Oliver' or 'Edwin' would present large frequency differences without being pertinent.

$$\text{Z score}(t_{ij}) = \left(rtf_{ij} - \overline{rtf_i}\right) \Big/ s_i \qquad (3)$$

Given a query text Q whose attribution is uncertain and a Text $A_j$ (written by the author $A_j$), we can calculate the difference in absolute value between the Z scores of texts Q and $A_j$. Then, the average is computed according to Equation 4, which is related to the Manhattan distance (with the insertion of the coefficient 1/m), where *m* indicates the number of terms selected and $t_{iAj}$ the *i*th term in Text $A_j$ (similarly for $t_{iQ}$ and Text Q).

$$\Delta(Q, A_j) = 1/m \cdot \sum_{i=1}^{m} \left| Z\ score\ (t_{iQ}) - Z\ score\ (t_{iA_j}) \right| \qquad (4)$$

In our experiments, the value of *m* varies from 50 to 800 (see Table 5). As *m* varies from one application to another, one cannot calibrate the returned distances. With Formula 4, the smallest value is 0 (the two texts have the same frequencies) but no upper limit can be defined. The lowest value of Delta indicates the probable authorship for texts Q and $A_j$.

As a new use of the Delta model, we propose estimating the probability that the computed distance indicates the real author. To achieve this, we suggest applying the softmin() function on all the distances, as expressed in Equation 5 where Prob($A_j$) reports the probability that Text Q was written by author $A_j$.

$$\text{Prob}(A_j) = e^{-c \cdot \Delta(Q, A_j)} \Big/ \sum_{i=1}^{k} e^{-c \cdot \Delta(Q, A_i)} \qquad (5)$$

The constant *c*, fixed at 20 in our experiments, is used to obtain a larger dispersion of the values returned of the Delta function.

| | With Matthew & Mark's Gospels | | Without Matthew & Mark's Gospels | |
|---|---|---|---|---|
| Features | Luke's Gospel | Acts | Luke's Gospel | Acts |
| 50 | 0.442† (0.945) | 0.654 (0.845) | 0.550 (0.997) | 0.604 (0.999) |
| 75 | 0.436† (0.896) | 0.670 (0.844) | 0.653 (0.990) | 0.615 (0.999) |
| 100 | 0.484† (0.762) | 0.690 (0.858) | 0.660 (0.993) | 0.681 (0.999) |
| 150 | 0.522† (0.828) | 0.710 (0.911) | 0.715 (0.987) | 0.666 (0.998) |
| 200 | 0.502† (0.879) | 0.724 (0.947) | 0.709 (0.972) | 0.699 (0.999) |
| 250 | 0.521† (0.851) | 0.727 (0.961) | 0.689 (0.971) | 0.705 (0.999) |
| 300 | 0.530† (0.817) | 0.728 (0.969) | 0.680 (0.966) | 0.695 (0.998) |
| 400 | 0.528† (0.830) | 0.747 (0.960) | 0.677 (0.946) | 0.694 (0.998) |
| 500 | 0.568† (0.677) | 0.764 (0.945) | 0.699 (0.933) | 0.704 (0.998) |
| 600 | 0.566† (0.747) | 0.743 (0.954) | 0.699 (0.938) | 0.728 (0.997) |
| 800 | 0.597† (0.747) | 0.794 (0.959) | 0.719 (0.864) | 0.756 (0.997) |

Table 5: Authorship attribution based on Burrows' Delta

To identify the real author of Luke and the Acts, the Delta model has been applied with the four Gospels, four Pauline letters, and the Book of Revelation. The returned values by Delta are shown in the second and third columns of Table 5. As depicted in this table (and as expected), the distance values increase when considering more terms. Second, the nearest to Luke is always Matthew's Gospel. This finding is indicated by a '†' after the Delta score. Figures in parentheses are the probabilities of a correct assignment (see Equation 5).

When looking for the nearest text to the Acts of the Apostles, the values of the third column in Table 5 correspond to Luke. One can see that the estimated probabilities of a correct assignment increase up to 300 terms then this value reaches a plateau.

The last two columns provide Delta values obtained when the Gospels of Matthew and Mark are removed. In this case the closest text to Luke is Acts. As mentioned in Section 3, the three synoptic gospels (Matthew, Mark, and Luke) contain many passages that are very similar. This common part has a clear impact on the computation of the Delta score. When the Gospels of Matthew and Mark are excluded, the resulting probability estimates produce a higher confidence that Luke is the author of both the Gospel and Acts. However, we must recall to the reader that such probability values must take them with a grain of salt (e.g. the number of texts used in this study is limited).

## 6. Authorship Verification

The Delta model tends to confirm common authorship for Luke-Acts. To supplement our analysis, a second model has been applied based on several different text representations. As a first strategy, the stylistic representation will be based on the *m* most frequent words, for *m* = 50, 75, 100, 150, 200, 300, 400, 500, 600, or 800. However, and following Koppel & Seidman (2018), each book is now subdivided into chunks of 5,000 tokens. Then, each pair of chunks generated from Luke-Acts (set denoted A) is compared to all other chunks extracted from the impostor group (set C). To achieve this, the four Pauline letters, John's Gospel and the Book of Revelation play the imitator role.

```
For a set of texts A = {a_1, ..., a_i, ..., a_j, .. } supposed to be written by A, a feature set,
        and a set of texts C = {c_1, c_2, ... c_v}
For each pair (a_i, a_j) {
    sim2(a_i, a_j) = 0
    Repeat r times {
        Randomly select 50% of the features set
        Compute sim(a_i, a_j)      # a first order similarity
        For all texts c_k  k=1, 2, ..., v:
            compute sim(a_i, c_k) and sim(a_j, c_k)
        If ( sim(a_i, a_j)*sim(a_i, a_j)  >  max{ sim(a_i, c_k) }  *  max{ sim(a_j, c_k) } )
            sim2(a_i, a_j) = sim2(a_i, a_j) + 1/r
        }
    }  # take the next pair of texts
  dec(a_i) = Aggregate { sim2(a_i, a_j) over all j }       #  e.g. the mean, the median, etc.
  if ( dec(a_i)  > δ )
      return "a_i was written by the proposed author"
      else return "a_i was written by an unknow author"
```

Table 6: Authorship verification algorithm (Koppel & Seidman, 2018)

The basic idea is to compute the similarity between two text representations extracted from the A set (denoted $a_i$ and $a_j$ in Table 6). One can select a similarity measure and compute the similarity value, indicated by sim($a_i$, $a_j$) (a first-order similarity). When the same author is behind two texts, this value should be high. Of course, one can use a distance measure instead of a similarity measure. In such cases, the interpretation and the decision rules must be reversed.

But to what does a high similarity value correspond? To provide an answer, the similarity score computed between $a_i$ and $a_j$ is compared with all *v* texts belonging to impostor set C. Only the largest similarity score with items in set C and $a_i$ and then $a_j$ is retained. Then, we multiply them. If this product is smaller than sim($a_i$, $a_j$) squared, one can conclude that sim($a_i$, $a_j$) presents a high score and provides evidence that $a_i$ and $a_j$ have been written by the same author. As shown in Table 6, this procedure is then repeated *r* times (e.g. *r* = 200, 500, or 1,000). To avoid repeating the same results over each iteration, only a fraction (in our case, 50%) of the entire feature set is employed. For example, when the whole sample contains 200 terms, only 100 are used to compare two texts.

After considering all pairs of chunks for a given $a_i$, one can conclude whether or not the text $a_i$ was really written by A. At this stage, the value sim2($a_i$, $a_j$) corresponds to a second-order similarity score with a value between zero and 1.0. To make this final decision, one needs to aggregate the second-order similarity values for each $a_i$ (denoted by dec($a_i$)). As a possible solution, one can compute the mean (as is the case in this study) or the median. Finally, if this aggregated value is higher than a predefined threshold $\delta$ (fixed at 80% in this study), one can consider that the text $a_i$ has been authored by A.

This ad hoc decision is binary. A more flexible decision could be used with a second threshold (denoted $\gamma$ and equal to 50%, for example). When dec($a_i$) is smaller than $\gamma$, the system has enough evidence to specify that $a_i$ was not written by A. Between the two thresholds ($\delta$ and $\gamma$), the answer is inconclusive, and the system returns "I don't know who is the real author".

1,000 iterations have been performed for each feature size in this study. To compare two text representations, the cosine similarity has been selected (see Equation 6). This choice is not grounded on a solid foundation; no formal argument can be put forward in favor of this or another similarity (or distance) function. The cosine corresponds to a well-known choice in information retrieval (Manning et al. 2008). As an advantage, the returned value is always between 0 (nothing in common) and 1.0 (identical texts). As an alternative, the Dice similarity function (see Equation 7) has the same advantage. Recommended by (Koppel & Seidman 2018), the MinMax function (Formula 8) has an overall good performance in a previous study. As a distance metric, the Manhattan (Equation 4) has the benefit of being simple to compute. As a variant, the Tanimoto (Formula 9) provides some form of normalisation while the Euclidian (Equation 10) distance corresponds to the classical concept of the distance between two points.

$$Sim_{Cosine}(A,B) = \sum_{i=1}^{m} a_i \cdot b_i \Big/ \sqrt{\sum_{i=1}^{m} a_i^2} \cdot \sqrt{\sum_{i=1}^{m} b_i^2} \tag{6}$$

$$Sim_{Dice}(A,B) = 2 \cdot \sum_{i=1}^{m} a_i \cdot b_i \Big/ \sum_{i=1}^{m} a_i^2 + \sum_{i=1}^{m} b_i^2 \tag{7}$$

$$Sim_{MinMax}(A,B) = \sum_{i=1}^{m} min(a_i, b_i) \Big/ \sum_{i=1}^{m} max(a_i, b_i) \tag{8}$$

$$Distance_{Tanimoto}(A,B) = \sum_{i=1}^{m}|a_i - b_i| \Big/ \sum_{i=1}^{m} max(|a_i|,|b_i|) \quad (9)$$

$$Distance_{Euclidian}(A,B) = \sqrt{\sum_{i=1}^{m}(a_i - b_i)^2} \quad (10)$$

The results of these experiments are reported in Table 7, in which the first column shows the number of stylistic features. For each similarity/distance measure, the method gives a vector of six values which corresponds, in order, to the three chunks extracted from Luke and the three extracted from the Acts. Instead of presenting this result vector, a mean over the six values is computed and given in the table. The maximum value is 1.0, indicating that each chunk is always found to be written by Luke for the 1,000 iterations.

Overall, the data depicted in Table 7 is clear: the model confirms a common authorship for all experiments. Moreover, when the number of employed features is less than 200 (presented in italics in the table), the results must be interpreted with caution. In such cases, the stylistic fingerprints of the author are represented by a relatively small number of characteristics that render the conclusion less reliable.

| | Similarity | | | Distance | | |
|---|---|---|---|---|---|---|
| Features | Cosine | MinMax | Dice | Manhattan | Tanimoto | Euclidian |
| 50 | *0.873** | *0.890** | *0.876** | *0.918* | *0.918* | *0.863** |
| 75 | *0.877** | *0.899** | *0.878** | *0.921* | *0.921* | *0.863** |
| 100 | *0.883** | *0.920* | *0.882* | *0.933* | *0.932* | *0.870** |
| 150 | *0.891** | *0.928* | *0.887* | *0.944* | *0.943* | *0.874** |
| 200 | 0.900 | 0.942 | 0.900 | 0.957 | 0.957 | 0.883 |
| 300 | 0.908 | 0.969 | 0.907 | 0.980 | 0.979 | 0.890 |
| 400 | 0.911 | 0.974 | 0.910 | 0.981 | 0.981 | 0.892 |
| 500 | 0.917 | 0.982 | 0.912 | 0.985 | 0.984 | 0.893 |
| 600 | 0.916 | 0.979 | 0.910 | 0.984 | 0.983 | 0.894 |
| 800 | 0.918 | 0.983 | 0.909 | 0.987 | 0.986 | 0.893 |

Table 7: Results with chunks of 5,000 tokens and different similarity/distance measures (1,000 iterations, random selection of 50% of the features)

In addition, when one of the components of the result vector is lower than the given threshold, an asterisk appears after the mean shown in Table 7. For example, when applying the cosine similarity with 50, 75, 100, and 150 features, the result vector has a first component that is lower than 0.8 (the δ threshold). A similar finding appears with the MinMax, Dice, and Euclidian functions. In all of those cases, only one of the six values is less than the threshold.

A deeper analysis of different attributions such as these reveals that the last fraction of Luke's Gospel is sometimes (and with a relatively small number of features) closely similar to the first part of John's Gospel. In other circumstances, such as the last portion of Acts (Paul travelling at sea), the system found some similarity with the beginning of John's Gospel.

Finally, to have a better understanding of the possible scores appearing in the result vector, the distribution of the values of the first component (extracted from the beginning of Luke) is displayed in Figure 1. To achieve this, 200 repetitions (over 1,000 iterations with a chunk size of 5,000 tokens) have been performed. The mean (989.05, standard deviation: 1.43) is represented by a solid line while the value used in Table 7 is signaled by the symbol 'Pt'. Overall, the distribution is clearly centered around its average without outliers.

As a second stylistic representation strategy, each book is decomposed into 3, 4, and 5 overlapping *n*-grams. For example, from the phrase "we give thanks", the following 4-grams are generated: {"we_g", "e_gi", "_giv", "give", …, "anks", } where "_" indicates a space. To reflect the author fingerprint, the 50, 75, 100, 150, 200, 300, 400, 500, 600, or 800 most frequent *n*-grams have been used to form the feature set. Based on the entire book, the results achieved with the Manhattan distance are reported in Table 8. As shown, this approach always returns the same answer: Luke and Acts were authored by the same person. The same experiments have been performed with the MinMax function (results appearing in the Appendix). Here, a similar conclusion can be drawn: Luke is the author of Luke and Acts.

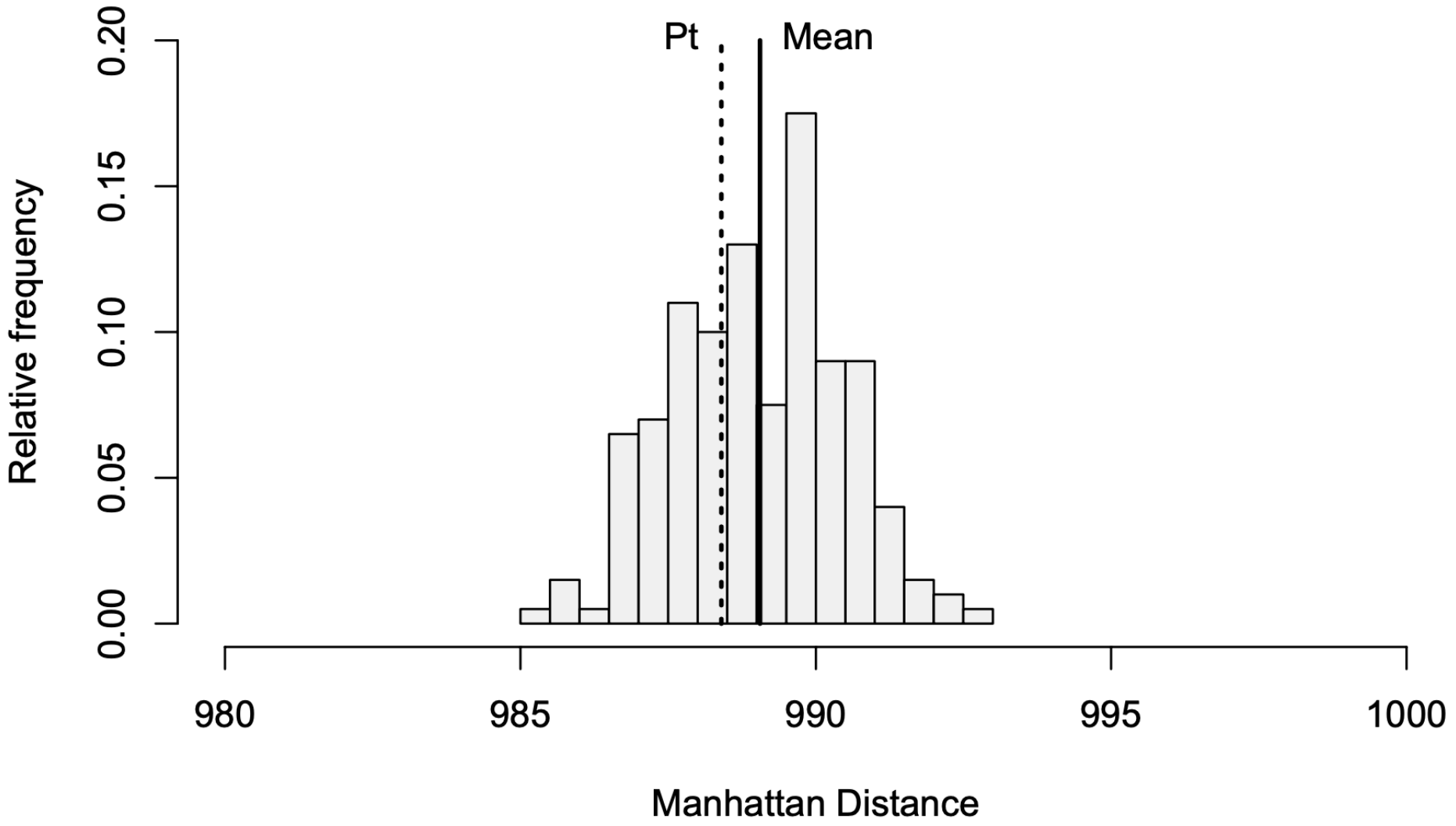

Figure 1: Histogram of the values returned by the verification algorithm for Luke's Gospel

A third set of experiments have been conducted based on truncation words denoted "trunc-n" for $n = 3, 4, 5$, and 6. In this case and for each book, only the first *n* letters of each word are kept. For example, from the phrase "Jesus provides us…", the following trunc-4 can be generated: {"jesu", "prov", "us"}. As a feature set, the 50, 75, 100, 150, 200, 300, 400, 500, 600, or 800 most frequent trunc-*n*, have been employed. The results depicted in Table 8 provide the same answer. This strategy always detects a common authorship. The table relating to the MinMax function in the Appendix confirms this conclusion.

As is the case for many other stylometric studies, several parameters must be fixed before performing an experiment. In the current case, the final conclusion is grounded on a solid foundation. Several variations of the parameters have been tested and always return the same conclusion. Moreover, and based on previous studies in the domain, the number of features must be 100 or more. This limit makes sense when working with the Greek language, which has a more complex morphology and a larger number of functional terms than in English.

| Terms | *n*-grams 3 / 4 / 5 | Trunc-*n* 3 / 4 / 5 / 6 |
|---|---|---|
| 50 | *0.760 / 0.282 / 0.766* | *0.997 / 1.0 / 1.0 / 1.0* |
| 75 | *0.811 / 0.860 / 0.987* | *1.0 / 1.0 / 1.0 / 1.0* |
| 100 | *0.843 / 0.988 / 0.998* | *0.999 / 1.0 / 1.0 / 1.0* |
| 150 | 0.999 / 0.999 / 1.0 | 1.0 / 1.0 / 1.0 / 1.0 |
| 200 | 1.0 / 1.0 / 1.0 | 1.0 / 1.0 / 1.0 / 1.0 |
| 300 | 1.0 / 1.0 / 1.0 | 1.0 / 1.0 / 1.0 / 1.0 |
| 400 | 1.0 / 1.0 / 1.0 | 1.0 / 1.0 / 1.0 / 1.0 |
| 500 | 1.0 / 1.0 / 1.0 | 1.0 / 1.0 / 1.0 / 1.0 |
| 600 | 1.0 / 1.0 / 1.0 | 1.0 / 1.0 / 1.0 / 1.0 |
| 800 | 1.0 / 1.0 / 1.0 | 1.0 / 1.0 / 1.0 / 1.0 |

Table 8: Results based on the Manhattan distance with *n*-grams and trunc-*n* representation (with a random selection of 50% of the features)

## 7. Stylistic Analysis

Based on considerations presented in Section 4, we can associate to each word (or lemma, bi-gram of words, etc.) a Z score indicating whether this term is employed more or less frequently by an author compared to the others (Muller 1992). To achieve this, the probability of observing a term, denoted *t*, selected randomly from the corpus is indicated by p(*t*). Based on the maximum likelihood principle, this probability would be estimated as p(t) = $tf_0 / n_0$ where $tf_0$ indicates the term frequency in the corpus and $n_0$ its size (in number of tokens) (Manning & Schütze 2001).

To verify whether this word is overused in a subcorpus of size $n_1$, its expected mean (estimated by $n_1 \cdot p(t)$) is compared to the observed occurrence frequency (denoted $tf_1$). Indeed, when randomly selecting $n_1$ times with replacement a word in this subcorpus, one can, in average, obtain $n_1 \cdot p(t)$ occurrences. Therefore, any large difference between $tf_1$ and $n_1 \cdot p(t)$ indicates a deviation from the expected behavior. To establish a more precise valuation of *large*, the variance of this binomial process (defined as $n_1 \cdot p(t) \cdot (1-p(t))$) can be taken into account. Equation 11 defines the final standardized Z score (or standard Gaussian distribution N(0,1)) for term *t*, using the partition between a given author (e.g. Luke) and the others (Matthew, Mark, John, 'John', and Paul).

$$\text{Z score}(t) = \left(tf_1 - (n_1 \cdot p(t))\right) \Big/ \sqrt{n_1 \cdot p(t) \cdot (1 - p(t))} \qquad (11)$$

For each term, we apply this procedure to compute its characteristic score according to the underlying text. Based on the Z score value, we then verify whether this term is used proportionally, with roughly the same frequency in both parts (Z score value close to 0). On the other hand, when a term is assigned a positive Z score larger than δ (e.g. 3), we consider it to be overused or belonging to the vocabulary characteristic of the target author.

For example, the word *Παῦλος* (Paul, noun, masc. sing. nom.) appears 55 times in Luke's writings, but only 9 occurrences appear with the other writers. Knowing that the subcorpus written by Luke contains 37,932 tokens while the entire collection encompasses 113,725 words, the expected number of *Παῦλος* in Luke must be 37,932 · ((55+9) / 113,725) = 33.65, a value lower than the observed number (55). The resulting Z score is 7.28 indicating an overused term under Luke's pen.

As a variant, one can consider that the occurrence frequencies must be smoothed to take account of possible unseen terms (Manning & Schütze 2001). As a first smoothing approach, Laplace suggests adding one to the occurrence frequency and likewise adding the vocabulary size to the denominator of $tf_0 / n_0$. This approach could then be generalized by using a λ parameter (Lidstone's law (1902)) defined between 0 and 1. The new probability estimate is provided by the following formula:

$$p(t) = {(tf_0 + \lambda)} \big/ {(n_0 + \lambda \cdot |V|)} \qquad (12)$$

in which $n_0$ indicates the size of the corpus (number of tokens), and |V| the number of distinct words in the corpus (or the vocabulary size). When applying this approach, we suggest to specify λ = 0.1 to avoid given to too large weight to infrequent terms (as applied in this study). On the other hand, when λ = 1, Laplace's smoothing is obtained. Applying this variant, the Z score associated to *Παῦλος* (Paul) is now 7.37.

| Luke | | | Matthew, Mark, John, Paul & 'John' | | |
|---|---|---|---|---|---|
| Z score | Frequency | Word | Z score | Frequency | Word |
| 12.31 | 160 | τε (and) | 6.13 | 438 | ἵνα (there) |
| 9.84 | 1,034 | δὲ (and) | 5.46 | 266 | λέγει (he says) |
| 7.51 | 123 | ἐγένετο (happened) | 5.10 | 377 | ἰησοῦς (Jesus) |
| 7.44 | 300 | εἶπεν (he said) | 4.96 | 633 | ἡ (the [femi.]) |
| 7.42 | 52 | ἄνδρες (men) | 4.54 | 463 | οὐ (not) |
| 7.37 | 55 | Παῦλος (Paul) | 4.53 | 473 | ἐκ (from) |
| 7.04 | 285 | πρὸς (to) | 4.40 | 1,866 | ὁ (the [masc.]) |
| 6.73 | 71 | κυρίου (of Lord) | 4.23 | 569 | γὰρ (indeed) |
| 6.27 | 30 | Παῦλον (to Paul) | 4.12 | 213 | ἐὰν (if) |
| 5.63 | 73 | αὐτούς (them) | 3.97 | 179 | ἀλλὰ (but) |

Table 9: Words overused by Luke (Gospel & Acts)
Matthew, Mark, John (Gospels), and 'John' (Revelation) & Paul (four Letters)

Table 9 reports other examples of words overused by either Luke or by the others authors (namely Matthew, Mark, John, 'John' and Paul). With Luke, one can see overused the conjunction 'and' (τε, δὲ, δέ), some prepositions (πρὸς (to), σὺν (with)), personal pronouns related to the third person (them (αὐτούς), him), and the verb 'to say' (εἶπεν). In addition, Luke overuses some nouns or names (Paul, men, Lord). On the other hand, other authors chose more frequently the name 'Jesus', negations, and some prepositions (from, for). Moreover, the nouns are usually more often overused as indicated by the overused of the definite article (ἡ, or ὁ).

As shown in Table 9, a translation is provided; one must take them with a grain of salt. Based on an single occurrence, a precise translation is rather difficult. For example from the conjunction *τε*, one can opt for the term 'and or 'furthermore'. As a more complex example, the term *ἐγένετο* is translated as 'he become' but a possible more precise translation could be 'come into a new state,' 'to have been born' or 'to have passed'.

| Luke | | | Matthew, Mark, John, Paul & 'John' | | |
|---|---|---|---|---|---|
| Z score | Freq. | Bigram of words | Z score | Freq. | Bigram of words |
| 9.60 | 68 | εἶπεν δὲ (and he said) | 4.71 | 239 | ὁ Ἰησοῦς ([the] Jesus) |
| 7.78 | 43 | εἶπεν πρὸς (he said to) | 3.88 | 160 | ἐκ τοῦ (from the) |
| 7.74 | 46 | πρὸς αὐτούς (to them) | 3.85 | 77 | καὶ λέγει (and he says) |
| 7.38 | 37 | ὁ Παῦλος ([the] Paul) | 3.76 | 70 | λέγει αὐτοῖς (says to them) |
| 7.18 | 35 | ἐγένετο δὲ (and he passed) | 3.45 | 75 | λέγει αὐτῷ (says to him) |
| 6.26 | 43 | τοῦ κυρίου (of the Lord) | 3.41 | 199 | καὶ ὁ (and the) |
| 5.93 | 24 | τὸν Παῦλον ([the] Paul) | 3.10 | 64 | ἀμὴν λέγω (Amen I say) |
| 5.67 | 34 | δὲ πρὸς (and to) | 3.07 | 44 | Ἰησοῦς καὶ (Jesus and) |
| 5.41 | 25 | δὲ αὐτῷ (and to him) | 3.04 | 88 | ἐκ τῶν (from the) |
| 5.38 | 29 | ὡς δὲ (and as) | 2.97 | 103 | καὶ ἡ (and the) |

Table 10: Bigrams of words overused by Luke (Gospel & Acts),
Matthew, Mark, John (Gospels), and 'John' (Revelation) & Paul (four Letters)

When considering overused bigram of words shown in Table 10, the context is slightly larger but for a precise translation, considering only a window of two words is not enough. Moreover, this table confirms the findings of the previous one.

First, Table 10 shows that Luke more frequently employs the past tense for the verb 'to say' (εἶπεν) while the other writers prefer another verb in the present tense (λέγει)[8]. Second, the names 'Paul' or 'Lord' occur more frequently in Lukan texts, while 'Jesus' (Ἰησοῦς) appears more often with the other authors.

Third, Luke's style is more oriented towards the action using more verbs, while both other authors tend to be more descriptive (more nouns). As the Greek morphology is richer than English, the definite article (the) appears under different forms as depicted in Table 10. For example, under Luke's pen, the genitive case of the definite article occurs frequently (τὸν, masc. sing., τῆς, femi. sing., or τοῦ, neut. sing.). This last form (τοῦ) also occurs in Matthew, Mark, John, 'John' & Paul's writings, though the nominative case is more frequently used for these authors (e.g. ὁ masc. sing., ἡ, femi. sing.).

At the lexical level, Table 9 and 10 show that the written style of an author can be characterized by higher frequencies of some functional terms (e.g. the, a, them, to, from, but, etc.). Some content words can also be useful when the same concept or idea can be expressed by different forms (e.g. to say or to tell; a boat, ship or vessel). In addition, some verbs could appear in distinct tenses according to different authors (e.g. present vs past tense). Third, the topics might reveal some additional stylistic markers or preferences (e.g. Paul, Jesus, men).

[8] As in English, one can use *to say*, *to tell*, *to speak*, or *to talk* depending on the context.

## 8. Conclusion

The question of common authorship between Luke's Gospel and the Acts of the Apostles is a recurring authorship verification problem. Previous studies have been conducted to corroborate or refute this hypothesis. Some of them are based on qualitative judgments about the style of the two books. Few investigations have considered the occurrence frequency differences of several words between the Gospel and Acts. To date, a definitive conclusion has not been reached (Burkett 2019; Gregory & Rowe 2010; Pervo 2010).

In this paper, the question has been examined using more recent authorship attribution strategies and author verification approaches. Based on Burrows' Delta (Burrows 2002) with eleven feature set sizes, this study always indicates common authorship of Luke's Gospel and the Acts. To confirm this common authorship, a verification model (Koppel & Seidman 2018) has been used with eight stylistic representations, six similarity (or distance) measures, and ten different feature sizes. This second strategy always confirms the same conclusion, namely that of common authorship. This study must be viewed as a first step in investigating Lukan authorship. One must acknowledge that the application of modern statistical modeling to Christian literature is a largely open domain (Pracht 2025).

The results also point to the central role played by a relatively large number of stylistic features. Grounding an authorship decision on a few words (see Section 3) is certainly not reliable. Our experiments tend to show that the number of features must be 100 or more for a reliable assessment. Moreover, it is good practice to consider different values for the underlying parameters (text representation strategy, feature size, distance measure, replication).

One must recognise the literary dependence of the synoptic gospels. Christian texts are subject to all kinds of textual variations introduced by copyists. This aspect could interfere with the truthful stylistic signal and renders the identification of the genuine writer harder. Studies on authorship attribution based on translated novels tend to demonstrate that the proposed approaches are still reliable, even in a slightly noisy context (Rybicki & Heydel, 2013; Rybicki 2025).

However, we agree that it is hard to establish the initial text or to determine whether a single author is really behind each work. Finally, as we are working with early Christian texts, the set of imitators is rather limited. It is difficult to obtain numerous documents that are written on the same topics, approximatively in the same time period, longer than 10,000 words, and by a single author (known with absolute certainty).

**Acknowledgments**

The author would like to thank the anonymous referees for their helpful suggestions and remarks.

## Appendix

| Mark | Matthew | Luke |
|---|---|---|
| 13,3 And as he sat upon the mount of Olives over against the temple, Peter and James and John and Andrew asked him privately,<br>13,4 Tell us, when shall these things be? and what shall be the sign when all these things shall be fulfilled? | 24,3 And as he sat upon the mount of Olives, the disciples came unto him privately, saying, Tell us, when shall these things be? and what shall be the sign of thy coming, and of the end of the world? | 21,7 And they asked him, saying, Master, but when shall these things be? and what sign will there be when these things shall come to pass? |
| 2,25 And he said unto them, Have ye never read what David did, when he had need, and was an hungred, he, and they that were with him? | 1,3 But he said unto them, Have ye not read what David did, when he was an hungred, and they that were with him; | 1,3 And Jesus answering them said, Have ye not read so much as this, what David did, when himself was an hungred, and they which were with him; |

Table A.1. Other examples of strong similarity between the synoptic gospels

| Matthew | Mark |
|---|---|
| 15,1 Then came to Jesus scribes and Pharisees, which were of Jerusalem, saying,<br><br><br><br><br>15,2 Why do thy disciples transgress the tradition of the elders? for they wash not their hands when they eat bread. | 7,1 Then came together unto him the Pharisees, and certain of the scribes, which came from Jerusalem.<br>7,2 And when they saw some of his disciples eat bread with defiled, that is to say, with unwashen, hands, they found fault<br>7,3 For the Pharisees, and all the Jews, except they wash their hands oft, eat not, holding the tradition of the elders.<br>7,4 And when they come from the market, except they wash, they eat not. And many other things there be, which they have received to hold, as the washing of cups, and pots, brasen vessels, and of tables.<br>7,5 Then the Pharisees and scribes asked him, Why walk not thy disciples according to the tradition of the elders, but eat bread with unwashen hands? |

Table A.2: Example of a clause added to Mark's Gospel

| Terms | *n*-grams<br>3 / 4 / 5 | Trunc-*n*<br>3 / 4 / 5 / 6 |
| --- | --- | --- |
| 50 | *0.990 / 0.999 / 0.993* | *1.0 / 1.0 / 0.991 / 0.991* |
| 75 | *1.0 / 1.0 / 0.999* | *1.0 / 1.0 / 0.998 / 0.998* |
| 100 | *0.999 / 1.0 / 1.0* | *1.0 / 1.0 / 0.999 / 0.998* |
| 150 | 1.0 / 1.0 / 1.0 | 1.0 / 1.0 / 1.0 / 1.0 |
| 200 | 1.0 / 1.0 / 1.0 | 1.0 / 1.0 / 1.0 / 1.0 |
| 300 | 1.0 / 1.0 / 1.0 | 1.0 / 1.0 / 1.0 / 1.0 |
| 400 | 1.0 / 1.0 / 1.0 | 1.0 / 1.0 / 1.0 / 1.0 |
| 500 | 1.0 / 1.0 / 1.0 | 1.0 / 1.0 / 1.0 / 1.0 |
| 600 | 1.0 / 1.0 / 1.0 | 1.0 / 1.0 / 1.0 / 1.0 |
| 800 | 1.0 / 1.0 / 1.0 | 1.0 / 1.0 / 1.0 / 1.0 |

Table A.3: Results based on the MinMax similarity with *n*-grams and trunc-*n* representation (with a random selection of 50% of the features)